\documentclass{CVM}
\usepackage{algorithmic}
\usepackage{amsmath}

\CVMsetup{
	type      = {Research Article},
	doi       = {s41095-0xx-xxxx-x},
	title     = {FCDFusion: a fast, low color deviation method for fusing visible and infrared image pairs},
	author    = {Hesong Li$^{1}$ and Ying Fu$^{1}$ \cor{}},
	runauthor = {H. Li, Y. Fu},
	abstract  = {
		Visible and infrared image fusion (VIF) aims to combine information from visible and infrared images into a single fused image. Previous VIF methods usually employ a color space transformation to keep the hue and saturation from the original visible image. However, for fast VIF methods, this operation accounts for the majority of the calculation and is the bottleneck preventing faster processing. In this paper, we propose a fast fusion method, FCDFusion, with little color deviation. It  preserves color information without color space transformations, by directly operating in RGB color space. It incorporates gamma correction at little extra cost, allowing color and contrast to be rapidly improved. We regard the fusion process as a scaling operation on 3D color vectors,  greatly simplifying the calculations. A theoretical analysis and experiments show that our method can achieve satisfactory results in only 7 FLOPs per pixel. Compared to state-of-the-art fast, color-preserving methods using HSV color space, our method provides higher contrast at only half of the computational cost. We further propose a new metric, color deviation, to measure the ability of a VIF method to preserve color. It is specifically designed for VIF tasks with color visible-light images, and overcomes deficiencies of existing VIF metrics used for this purpose. Our code is available at \url{https://github.com/HeasonLee/FCDFusion}.
	},
	keywords  = {infrared images, visible and infrared image fusion; gamma correction; real-time display; color metrics; color deviation},
	copyright = {The Author(s) 2022.},
}

\begin{document}
	
	\maketitle
	
	\begin{figure}[b] \vskip -4mm
		\small\renewcommand\arraystretch{1.3}
		\vskip 3mm
		\begin{tabular}{p{80.5mm}} \toprule\\ \end{tabular}
		\vskip -4.5mm \noindent \setlength{\tabcolsep}{1pt}
		\begin{tabular}{p{3.5mm}p{80mm}}
			$1\quad $ & School of Computer Science and Technology, Beijing Institute of Technology, Beijing 100081, China. E-mail: H. Li, 3120211055@bit.edu.cn; Y. Fu, fuying@bit.edu.cn \cor{}.\\
			&\hspace{-5mm} Manuscript received: 2022-01-01; accepted: 2022-01-01\vspace{-2mm}
		\end{tabular} \vspace {-3mm}
	\end{figure}
	\graphicspath{{figures/}}
	
	\section{Introduction}\label{sec:introduction}
	
	Visible and infrared image fusion (VIF) aims to combine information from visible and infrared images into a single fused image. In dark or dazzling extreme environments, an image taken by a visible camera may be unclear, but an infrared camera can usually obtain  clear object outlines due to differences in temperature. A VIF output image has both color and temperature information,  making objects in the output image clearer and more distinguishable. Therefore, VIF technology has been an active research field for many years, and has abundant applications \cite{survey} in many fields such as object detection \cite{detection,detect3,detect1,detect2}, tracking \cite{tracking,track1,track2}, recognition \cite{recog1,recog3,recog2}, surveillance \cite{surveil3,surveil1,surveil2}, color vision \cite{color-vision1}, and remote sensing \cite{remote1,remote2}.
	
	Several fast VIF methods simply average two input images in different color spaces \cite{IHS}, taking no more than 20 FLOPs for each pixel pair. Their main goal is to fuse video at high-speed for real-time display, so they trade off quality of results for speed. When putting image quality   first, rather than speed, other methods are used; representative ones being based on domain transformation~ \cite{PCA,LP,wavelet,NC,MSVD,LatLRR}, deep learning~\cite{DL2019,CNN,CUFD,PIAFusion,SeAFusion}, or hybrid methods ~\cite{hybrid2,hybrid3,MST,hybrid1,luo2022,yin2022}. These methods provide impressive results but involve much more complex calculation, often taking over 1000 times as many FLOPs as simple averaging methods.
	
	Previous VIF methods usually employ a color space transformation to keep the hue and saturation from the original visible image. However, for fast VIF methods, this operation accounts for the main part of the calculation, and is the bottleneck preventing improved processing speed. In this paper, we propose a fast fusion method, FCDFusion, with little color deviation; it can preserve color information without color space transformations. It directly operates in RGB color space and  embeds gamma correction with little extra computation, so both color and contrast can be quickly improved. We regard the fusion process as a scaling operation on 3D color vectors, which  greatly simplifies the calculations. Theoretical analysis and experimental results show that our method can achieve satisfactory results using only 7 FLOPs per pixel. Compared to state-of-the-art fast, color-preserving methods using HSV color space, our method provides higher contrast in half the number of FLOPs.
	
	We further observe that existing evaluation metrics for the VIF task mainly consider the sharpness of the fused image (via entropy \cite{EN}, average gradient \cite{AG}, etc.) and its overall structural similarity to the original images (via structural similarity index measure \cite{SSIM}, root mean squared error \cite{PSNR}, etc.), and ignore the color information. Thus, some methods achieve high scores by changing the hue and saturation of the original color to improve the contrast. In this work, we propose a new metric called color deviation to measure how well a  VIF method preserves colour. It is specifically designed for VIF tasks using color visible images and aims to remove the deficiencies of existing VIF metrics by assessing color information.
	
The main contributions of this work are thus:	
	\begin{itemize}
		\item  a simple and effective VIF method, FCDFusion, which can preserve color and improve contrast in only 7 FLOPs per pixel,
		\item a metric, color deviation, specifically designed to measure the ability of a VIF method, using color visible images,  to preserve color,
		\item a theoretical analysis, based on color deviation, of the color-preserving abilities of our method and averaging methods in RGB, YIQ, and HSV color spaces; these results are also verified experimentally.
	\end{itemize}
	
	\section{Related work} \label{sec:Relatedwork}
	In this section, we first review  existing VIF methods and their strategies for preserving color. Then we give a brief introduction to existing metrics used to evaluate VIF results.
	
	\subsection{VIF methods}
	\subsubsection{Problems with simple averaging}
	A simple VIF method working in RGB color space is to average the red, green, and blue components of a visible image with an infrared image separately. Unfortunately, there are two problems to be solved.
	Firstly, one of the input images usually contains more details, while the other is blurred. A simple averaging operation cannot deliberately select the image containing more detail, and therefore, after the averaging operation, the original clear details become blurred. 
	Secondly, because the infrared image contains only one channel (monochromatic), averaging the red, green, and blue components of the visible image with the same gray value will cause a decline in saturation, resulting in poor appearance.	
	Existing VIF methods have developed strategies to solve the above two problems, to help them keep details and colors in the fused image.
	
\subsubsection{Detail-preserving methods}
	To selectively keep  details from both input images, various more complex fusion methods have been proposed.
	
	Some  transform the two input images into a new domain to extract principal (low-frequency) parts and detail (high-frequency) parts from the input images. Typical domain transformations include principal component analysis \cite{PCA}, Laplacian pyramids \cite{LP}, wavelet transforms \cite{wavelet}, contourlet transforms \cite{NC}, multi-resolution singular value decomposition \cite{MSVD} and latent low-rank representation \cite{LatLRR}. In these methods, the two parts of the output image are fused in different ways. The high-frequency part of the output image is usually the sum or the maximum value of the two input images, while the low-frequency part is usually formed by averaging. Such processing aims to preserve the details of the two input images as well as possible. 
	
	To achieve better overall quality and avoid the disadvantages of a single method, some hybrid methods \cite{hybrid2,hybrid3,MST,hybrid1} mix the results of multiple methods to get the best effects in the output images. 
	
	Moreover, with the rapid development of deep learning, some deep learning-based methods \cite{DL2019,CNN} are also emerging. These methods take some evaluation metrics as training objectives and use various artificial neural network models for parameter training to achieve better fusion results.
	
	Recently, most methods combine neural networks with other techniques to improve the fusion effect. Inspired by domain transformation methods, Luo \emph{et al.} \cite{luo2022} use an $\ell1$-$\ell0$  decomposition model to obtain the base and detail layers before fusion, and employ Laplacian and Gaussian pyramids to decompose the detail layers and decision map obtained by a convolutional neural network. To fully preserve  visual details, Yin \emph{et al.} \cite{yin2022} employ a weighted mean curvature-based multiscale transform fusion scheme which can effectively suppress noise and keep valuable details. Xu \emph{et al.} \cite{CUFD} use convolutional neural networks as encoders to extract multi-level features from the input image pair, and then use a decoder to fuse these features. Tang \emph{et al.} design an illumination-aware sub-network in PIAFusion \cite{PIAFusion} to estimate the illumination distribution and calculate the illumination probability, then use the illumination probability to construct an illumination-aware loss to guide training of the fusion network. It thus performs well on target maintenance and texture preservation in areas with different illumination. In SeAFusion \cite{SeAFusion}, Tang \emph{et al.} cascade an image fusion module and a semantic segmentation module, using  semantic loss to guide the flow of high-level semantic information   back to the image fusion module,  effectively boosting the performance of high-level vision tasks on fused images, such as semantic segmentation and object detection.
	
	The above methods trade off speed for better quality of fused results. Although the SeAFusion \cite{SeAFusion} model is simplified to improve its speed, it still requires more calculation than  early deep learning-based methods like CNN \cite{CNN}, and about 10,000 times that of  simple averaging methods \cite{IHS}. Here, we propose a simple and effective VIF method to achieve high quality fusion results with extremely low computational requirements, akin to those of simple averaging methods.
	
	\begin{figure}[t!]
		\centering
		\includegraphics[width=\linewidth]{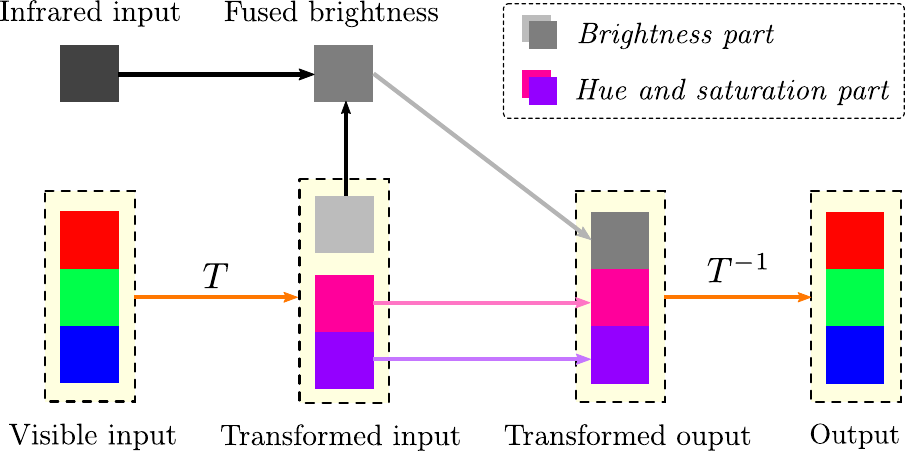}
		\caption{Color-preservation strategy of VIF methods using a color space transformation $T$.}
		\label{fig:colorFrame}
	\end{figure}

\subsubsection{Color-preserving methods}
	To keep the hue and saturation of the original visible image, VIF methods can be modified to operate in another color space in which one component represents brightness, and two other components represent hue and saturation~\cite{IHS}. Only the brightness component of the two images is fused, and the hue and  saturation of the original color image is retained. Such a color-preservation strategy is also commonly used in other image enhancement methods operating on color images \cite{enhance,enhance2}. As Fig.~\ref{fig:colorFrame} illustrates, color-preserving VIF methods first use a color space transformation $T$ to divide the visible image into hue and saturation components and a brightness component. They then fuse the brightness component with the infrared image using some particular fusion strategy, and finally, they use the inverse transformation $T^{-1}$ to transform the result back to RGB color space. 
	
	Two commonly used color spaces are YIQ and HSV \cite{YIQ,HSV}. When using YIQ color space, both $T$ and $T^{-1}$ are $3\times3$ matrices, so the VIF method needs two matrix multiplications. When using HSV color space,  saturation is better preserved but the conversion is more complex and requires more time.
	
	Some recent methods \cite{PIAFusion,SeAFusion} use YUV (also called YCrCb) \cite{YUV} color space, which is similar to YIQ color space. The transformation from RGB  to YUV  needs a 3$\times$3 matrix multiplication and two scalar additions.
	
	Color space transformations can help VIF methods keep the hue and saturation from the original visible image. But for fast VIF methods, transformations account for the majority of the calculation and act as a bottleneck to improving processing speed. Our method proposed in this paper directly operates in RGB color space and uses vector scaling instead of color space transformations.
	
	\subsection{VIF metrics}\label{sec:metrics}
	Numerous metrics used to evaluate image quality can also be used to evaluate the quality of fusion results. None of the proposed metrics is universally better than all others; they are complementary. Typical metrics used in VIF result evaluation \cite{metrics} include 
	information theory-based metrics (e.g.\ cross-entropy (CE) \cite{CE}, entropy (EN) \cite{EN}, mutual information (MI) \cite{MI}, and peak signal-to-noise ration (PSNR) \cite{PSNR}), 
	structural similarity-based metrics (e.g.\ structural similarity index measure (SSIM) \cite{SSIM}, and root mean squared error (RMSE) \cite{PSNR}), 
	image feature-based metrics (e.g.\ average gradient (AG) \cite{AG}, edge intensity (EI) \cite{EI}, standard deviation (SD) \cite{SD}, spatial frequency (SF) \cite{SF}, 
	and gradient-based fusion performance ($Q^\text{AB/F}$) \cite{QABF})  and human perception inspired metrics (e.g.\  the Chen-Blum metric ($Q_\text{CB}$) \cite{QCB}, and the Chen-Varshney metric ($Q_\text{CV}$) \cite{QCV}). 
	
		\begin{figure}[t!]
		\centering
		\includegraphics[width=.6\linewidth]{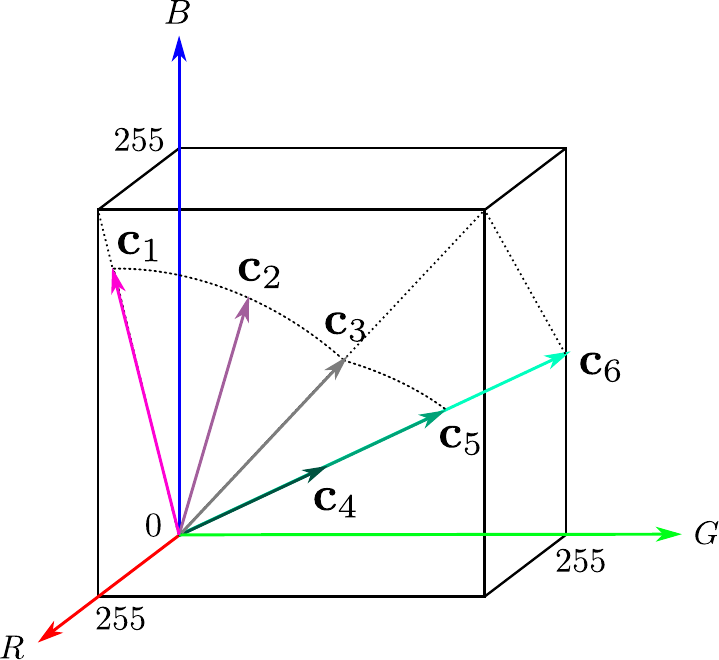}
		\caption{Color vectors in RGB color space.}
		\label{fig:colorVector}
	\end{figure}

	These existing metrics either measure the overall contrast of the output image or its similarity to each of the two original input images. 
	However, for VIF tasks with colored visible images, color information and brightness information of the image are usually processed separately, so it is necessary to design a metric that specifically measures color changes. In this work, we propose a new metric. color deviation, to measure the color preservation ability of a VIF method.

		\begin{figure*}
		\centering
		\includegraphics[width= \linewidth]{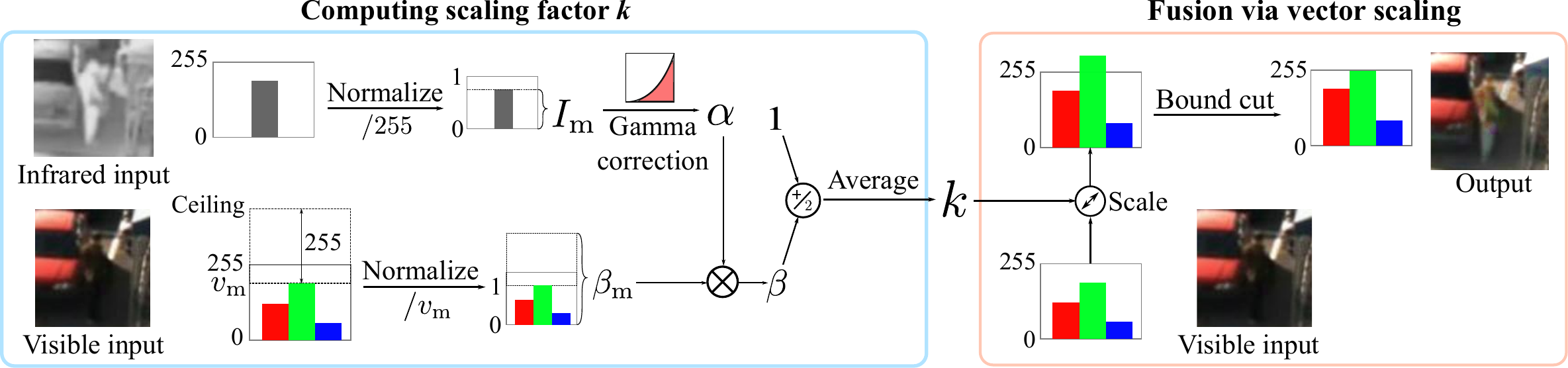}
		\caption{FCDFusion Framework. Gamma correction uses $\gamma=2$. Scaling multiplies each input RGB component by the same factor $k$.}
		\label{fig:frame}
	\end{figure*}

	\section{FCDFusion method}
	In this section, we first formulate the VIF problem and  motivate our method. Then we introduce our FCDFusion fast fusion method with low color deviation  for high-speed  VIF tasks demanding color preservation. It  can greatly reduce the computational load while ensuring good contrast and color quality in the fused image.

	\subsection{Formulation and motivation}
	In the visible and infrared image fusion (VIF) problem,  two images of the same scene are taken by a visible camera and an infrared camera; we assume they have been aligned by a registration algorithm, so that each has the same size and  each object has the same position  in both images. For pixel-level VIF methods, the goal is to calculate the fused pixel color $\mathbf{c}_\text{f}$ by using the visible pixel color $\mathbf{c}_\text{v}$ and the infrared pixel color $\mathbf{c}_\text{i}$ for each pixel pair across the two images. $\mathbf{c}_\text{f}$, $\mathbf{c}_\text{v}$, and $\mathbf{c}_\text{i}$ are color vectors in RGB color space with three elements:
	\begin{equation}
		\mathbf{c}_\text{f}=
		\left[\begin{array}{c}
			r_\text{f}\\g_\text{f}\\b_\text{f}
		\end{array}\right],\quad
		\mathbf{c}_\text{v}=
		\left[\begin{array}{c}
			r_\text{v}\\g_\text{v}\\b_\text{v}
		\end{array}\right],\quad
		\mathbf{c}_\text{i}=
		\left[\begin{array}{c}
			v_\text{i}\\v_\text{i}\\v_\text{i}
		\end{array}\right],
	\end{equation}
	where $r_\text{f}, g_\text{f}, b_\text{f}, r_\text{v}, g_\text{v}, b_\text{v}$, and $v_\text{i}$ are integers, usually in the range [0,255].
	
	Because the three elements of the infrared color $\mathbf{c}_\text{i}$ have the same value $v_\text{i}$ (it is monochromatic), fusing the red, green and blue components of the visible image with the same gray value may cause a decline in saturation, giving the output image a poor appearance. Thus, existing color-preserving methods \cite{IHS} introduce a  color transformation to a new color space such as YIQ or HSV to improve the color quality of the output image (see Fig.~\ref{fig:colorFrame}). They first transform the RGB color of the visible image into hue and saturation components and a brightness component, then fuse the brightness component with the infrared image, and finally use inverse transformation to transform the result back to RGB color space. The majority of the computation in this process lie in the color space transformation and the inverse transformation, while the fusion process is usually a simple averaging.
	
	However, if we consider the RGB color as a vector in RGB color space, this process can be greatly simplified: the brightness is the length of the vector, and the hue and saturation are represented by the direction of the vector.  See, for example,  Fig.~\ref{fig:colorVector}. The four color vectors $\mathbf{c}_1$, $\mathbf{c}_2$, $\mathbf{c}_3$, and $\mathbf{c}_5$ have the same lengths but different directions, so have the same brightness and different hues and saturations. 
	The closer the direction is to the main diagonal of the color space cube, the lower its saturation, and the closer it is to gray. So although the three color $\mathbf{c}_1$, $\mathbf{c}_2$, and $\mathbf{c}_3$ have the same hue, $\mathbf{c}_3$ is completely gray while $\mathbf{c}_1$ has the highest saturation.
	Another aspect, also shown in Fig.~\ref{fig:colorVector}, is illustrated by the three color vectors $\mathbf{c}_4$, $\mathbf{c}_5$, and $\mathbf{c}_6$, which have the same direction but different lengths. They thus have the same hue and saturation but different brightnesses. Therefore, a color-preserving fusion process can be considered to be one which simultaneously multiplies the red, green, and blue values by a common scaling factor $k$. In our method, $k$ depends on both input images.
	
	\subsection{Method}
	The framework of the proposed method is shown in Fig.~\ref{fig:frame}. We first compute the scaling factor $k$ using the input pixel pair, then fuse the pixel pair by scaling the visible input color vector by this factor $k$. We proceed in stages to determine $k$.
	
\subsubsection{Scaling ratio}
	The output color can be seen as a scaled version of the original visible color vector $\mathbf{c}_\text{v}$, where the scaling ratio $\alpha$ depends on the magnitude of the infrared value $v_\text{i}$:
	\begin{equation}\label{alpha}
		\alpha=\left(\frac{v_\text{i}}{255}\right)^\gamma\in{[0,1]},
	\end{equation}
	where the infrared input value $v_\text{i}$ is normalised to [0,1], and the exponent $\gamma$ is used to improve the contrast of the infrared image and reduce noise. 
	
	Gamma correction uses an exponential function to adjust the relative brightness of an image. Using $\gamma>1$ enhances  contrast in the bright parts of the picture and reduces contrast in the dark parts \cite{gamma}. There is always noise in dark areas due to the lower signal-to-noise ratio, and reducing the contrast in dark areas acts to suppress this noise. In the proposed method we set $\gamma=2$, allowing generic exponentiation to be replaced by a simpler and faster single multiplication in this case.
	
	The scaling ratio $\alpha$ represents the relative value of $k$. $\alpha=1$ means $k$ should take its maximal value $\beta_\text{m}$, to maximize the length of $\mathbf{c}_v$; $\alpha=0$ means $k$ should take its minimal value 0, making $\mathbf{c}_\text{v}$  black.
	
\subsubsection{Maximal scaling factor} 
	To simplify the calculation, we take the brightness of the color $\mathbf{c}_\text{v}$ to be approximately given by the maximum value of all three components. However, as $v_\text{m}$ appears in a denominator later, we prevent division by zero by ensuring $v_\text{m}\ge 1$:
	\begin{equation}
		v_\text{m}=\max(r_\text{v},g_\text{v},b_\text{v},1).
	\end{equation}
	
	To ensure that all color vectors have a chance to be enlarged, we define the ceiling of the scaled vector to have value $v_\text{m}+255$, so the maximal scaling factor is:
	\begin{equation}\label{eq:betam}
		\beta_\text{m}=\frac{v_\text{m}+255}{v_\text{m}}.
	\end{equation}
Thus, with a maximal scaling ratio $\alpha=1$, the color component with greatest value is scaled to the ceiling $v_\text{m}+255$; the other two components are also scaled by the same factor $\beta_\text{m}$.
	
\subsubsection{Scaling factor}
	The initial scaling factor $\beta$ is now obtained by multiplying the maximum scaling factor $\beta_\text{m}$ by the scaling ratio $\alpha$:
	\begin{equation}\label{eq:beta}
		\beta=\alpha \beta_\text{m}.
	\end{equation}
	A scaling operation directly using the factor $\beta$ is too severe: the brightness of the output image differs too much from that of the visible image. Averaging this value with the original visible image  reduces this difference:
	\begin{equation}\label{eq:cs}
		\mathbf{c}_\text{s}=
		\frac{\beta \mathbf{c}_\text{v}+\mathbf{c}_\text{v}}{2}
		=k \mathbf{c}_\text{v},
	\end{equation}
	where $\mathbf{c}_\text{s}$ is the scaled color vector after  averaging with the original color vector $\mathbf{c}_\text{v}$.
	Combining Eqs.~(\ref{eq:betam}--\ref{eq:cs}), we  obtain the final scaling factor $k$ as:
	\begin{equation}
		k=\frac{\alpha \beta_\text{m}+1}{2}=\frac{\alpha(v_\text{m}+255)/2}{v_\text{m}}+0.5\label{k},
	\end{equation}
	where the calculation of $(v_\text{m}+255)$ divided by 2 can be quickly performed by a right-shift operation on the integer $(v_\text{m}+255)$. 
	
\subsubsection{Fusion via vector scaling}
	The proposed method fuses the pixel pair by scaling the visible input color vector by the computed scaling factor $k$. It changes the length of the vector but keeps its direction. Therefore, while the visible image adds brightness information from the infrared image, its color information is preserved.
	
	However, after the scaling by $k$, some component of the color $\mathbf{c}_\text{s}$ may be greater than 255. The output color $\mathbf{c}_\text{f}$ is finally obtained by limiting the three channels of $\mathbf{c}_\text{s}$ to [0,255]:
	\begin{equation}\label{eq:cf}
		\mathbf{c}_\text{f}=
		\left[\begin{array}{c}
			r_\text{f}\\g_\text{f}\\b_\text{f}
		\end{array}\right]
		=
		\left[\begin{array}{c}
			\min(k r_\text{v},255)\\
			\min(k g_\text{v},255)\\
			\min(k b_\text{v},255)
		\end{array}\right].
	\end{equation}

The overall computation is presented in Algorithm~\ref{alg1}.
	
The analysis of arithmetic operations given in Tab.~\ref{tab:mult} shows that simple averaging methods only use simple integer operations and no floating point operations during fusion. In fast, color-preserving methods like YIQ-AVG and HSV-AVG, color space transformations account for the majority of floating point operations. Our method uses vector scaling (Eq.~\ref{eq:cf}) instead of color space transformations, so is faster. Further discussion is provided in Sec.~\ref{sec:results}.
	
	\begin{table*}[t!]
		\caption{FLOPs required by various fusion methods, per fused pixel, and their color-preservation performance. Columns 2--4 give the FLOPs for the 3 main stages: transformation from RGB color space, fusion, and inverse color space transformation.} 
		\centering
		\begin{tabular}{l|ccc|r|l}
\hline
\textbf{Method}  & \textbf{From RGB} & \textbf{Fusion} & \textbf{To RGB} & \textbf{Total} & \textbf{Color preservation} \\ \hline
RGB-AVG          & 0                 & 0               & 0               & 0              & poor                         \\
YIQ-AVG          & 9                 & 0               & 9               & 18             & intermediate                      \\
HSV-AVG          & 6                 & 0               & 8               & 14             & good                        \\
FCDFusion (ours) & 0                 & 7               & 0               & 7              & good                        \\ \hline
\end{tabular}\label{tab:mult}
	\end{table*}
	
	\begin{algorithm}[t]
		\begin{algorithmic}
			\STATE \textbf{Input:} RGB color $\mathbf{c}_\text{v}$ of visible pixel and corresponding value $v_\text{i}$ of infrared pixel.
			\STATE \textbf{Output:} RGB color $\mathbf{c}_\text{f}$ of fused pixel.
			\STATE $\alpha \longleftarrow v_\text{i}/255.0;$
			\STATE $\alpha \longleftarrow \alpha \times \alpha;$
			\STATE $v_\text{m} \longleftarrow 1;$
			\FOR {each RGB colour component $c_\text{v}$ of color $\mathbf{c}_\text{v}$}
			\IF {$c_\text{v}>v_\text{m}$}
			\STATE $v_\text{m} \longleftarrow c_\text{v};$
			\ENDIF
			\ENDFOR
			\STATE $k=(v_\text{m}+255)>>1;$
			\STATE $k \longleftarrow k\times\alpha/v_\text{m}+0.5;$
			\FOR {each RGB component $c_\text{v}$ of color $\mathbf{c}_\text{v}$, and\\ corresponding
				 component $c_\text{f}$ of color $\mathbf{c}_\text{f}$}
			\STATE $c_\text{f} \longleftarrow k\times c_\text{v};$
			\IF {$c_\text{f}>255$}
			\STATE $c_\text{f} \longleftarrow 255;$
			\ENDIF
			\ENDFOR
		\end{algorithmic}
		\caption{FCDFusion method}\label{alg1}
	\end{algorithm}
		\begin{figure*}
		\centering
		\includegraphics[width=.9\linewidth]{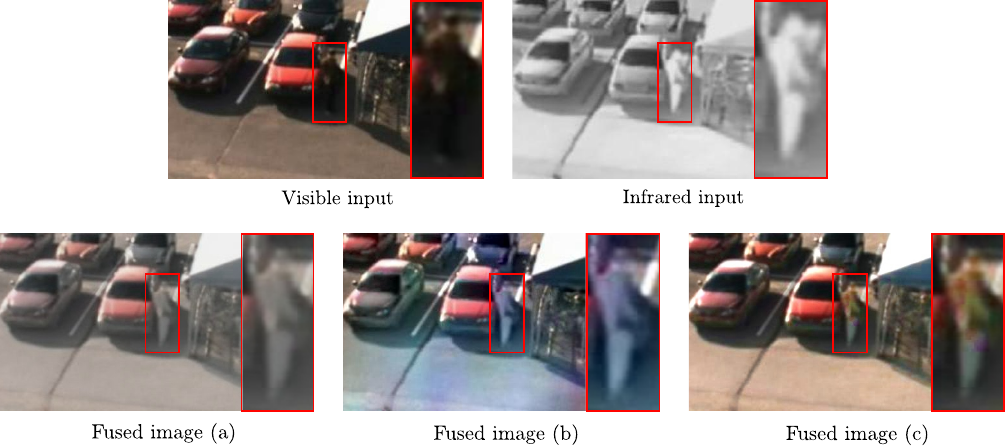}
		\caption{Color comparison of three fused images (a), (b), and (c), obtained by RGB-AVG, MST-SR, and our method, respectively. Only (c) presents a good visual effect and retains color information.}
		\label{fig:CDMotivation}
	\end{figure*}

	\section{Color deviation metric}
	In this section, we propose a new metric, \emph{color deviation} (CD), to measure the color-preservation ability of a VIF method. It is specifically designed for VIF tasks with color visible images and is intended to avoid the deficiencies of existing VIF metrics used to evaluate color preservation. We first motivate and define color deviation, and then analyze the color preservation ability of several fast VIF methods using this metric.

	\subsection{Motivation and definition}\label{sec:CDMotivation}
	Fig.~\ref{fig:CDMotivation} illustrates the deficiencies which can result when using existing VIF metrics. From both perspectives of clarity and color preservation, Fig.~\ref{fig:CDMotivation}(c) provides a superior  fused image to Figs.~\ref{fig:CDMotivation}(a,b) for the same two input images. The man in the car's shadow is enhanced in Fig.~\ref{fig:CDMotivation}(c), which also shows that the man wears a brown or orange coat. This color information is lost in Figs.~\ref{fig:CDMotivation}(a,b) by the changes in hue and saturation. However, when evaluated by existing metrics, which consider similarity  or contrast, Fig.~\ref{fig:CDMotivation}(c) is not considered  to be the best result. The main reason is that the existing metrics are defined specifically for monochrome images.
	
	Contrast-related metrics (such as CE \cite{CE}, EN \cite{EN}, MI \cite{MI}, AG \cite{AG}, EI \cite{EI}, SD \cite{SD}, SF \cite{SF}, $Q^\text{AB/F}$ \cite{QABF}, $Q_\text{CB}$ \cite{QCB} and $Q_\text{CB}$ \cite{QCV}) measure the whole contrast of the fused image, including contrasts in hue, saturation and brightness. Therefore, a fused image can obtain a higher score by increasing the contrast in hue and saturation channels, which may lead to large differences in hue and saturation between the visible image and the fused image. Using such  metrics, Fig.~\ref{fig:CDMotivation}(b) generally has the highest score of the three output images in Fig.~\ref{fig:CDMotivation}. Instead, to preserve color information from the visible image, hue and saturation need to be unchanged during fusion.
	
	Similarity-related metrics (such as PSNR \cite{PSNR}, SSIM \cite{SSIM} and RMSE \cite{PSNR}) compare the fused image to both  input images. Since the infrared image is monochromatic and it is given the same importance  as  the visible image, a fused image with saturation between the two input images is more likely to obtain a higher score. Thus, evaluated by such metrics, Fig.~\ref{fig:CDMotivation}(a) has the highest score among the three output images in Fig.~\ref{fig:CDMotivation}. In fact, only the visible image has hue and saturation information, so the similarity of hue and saturation should be compared to the visible image individually.
	
	To avoid these problems, in VIF tasks with color visible images, color information and brightness information should be measured respectively. 
	Brightness information (\emph{i.e.}, the brightness similarity of the two input images and the fused image, and the brightness contrast of the fused image itself) can be assessed by existing metrics after changing the visible image and fused image into grayscale images.	
	For color information, the color deviation (CD) between the visible input color and fused color can be measured by the angle between the two color vectors $\mathbf{c}_\text{f}$ and $\mathbf{c}_\text{v}$:
	\begin{equation}
		\text{CD}(\mathbf{c}_\text{v},\mathbf{c}_\text{f})=
		\arccos{\frac{\mathbf{c}_\text{v}\cdot \mathbf{c}_\text{f}}
			{\left|\mathbf{c}_\text{v}\right| \left|\mathbf{c}_\text{f}\right|}},
	\end{equation}
with arccos returning its principal value in the range $[0,\pi]$.
	A small color deviation value indicates that hue and saturation are well preserved from the visible input color. When angles are small, the differences from their cosine values are not obvious, so we use angles instead of their cosine values.
	
	To assess the overall color preservation ability of a VIF method for an input image pair, the average color deviation over all  pixels is used. 
	
	\subsection{Color deviation of existing fast methods}
	
	\begin{figure}
		\centering
		\includegraphics[width=.6 \linewidth]{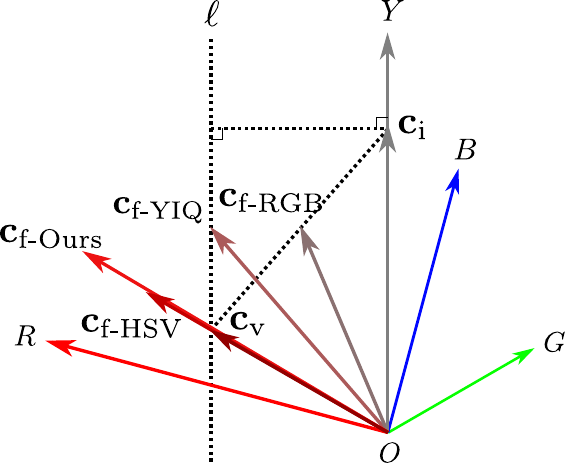}
		\caption{Directional comparison of input color vectors (\emph{i.e.}, $\mathbf{c}_\text{v}$ and $\mathbf{c}_\text{i}$)  and fused color vectors (\emph{i.e.}, $\mathbf{c}_\text{f-RGB}$, $\mathbf{c}_\text{f-YIQ}$, $\mathbf{c}_\text{f-HSV}$, and $\mathbf{c}_\text{f-Ours}$).}
		\label{fig:CD}
	\end{figure}
	
	Here, we analyze the color-preserving ability of four existing fast VIF methods using color deviation. In Fig.~\ref{fig:CD}, the $Y$ axis of YIQ color space is the diagonal of RGB color space, on which each color has the same red, green and blue value. So the infrared color $\mathbf{c}_\text{i}$ is along on the $Y$ axis.
	
	The VIF averaging method in RGB color space simply averages $\mathbf{c}_\text{v}$ with $\mathbf{c}_\text{i}$, so the fused vector $\mathbf{c}_\text{f-RGB}$ is the midpoint of the line segment connecting $\mathbf{c}_\text{v}$ and $\mathbf{c}_\text{i}$. As  Fig.~\ref{fig:CD} shows, this averaging operation causes a large angle between $\mathbf{c}_\text{v}$ and $\mathbf{c}_\text{f-RGB}$.
	
	The VIF averaging method in YIQ color space averages $\mathbf{c}_\text{v}$ with $\mathbf{c}_\text{i}$ only in the $Y$ channel. As  Fig.~\ref{fig:CD} indicates, doing so moves $\mathbf{c}_\text{v}$ along a straight line $\ell$ which is parallel to the axis $Y$. More specifically, the fused vector $\mathbf{c}_\text{f-YIQ}$ is the midpoint of the line segment connecting $\mathbf{c}_\text{v}$ and $\mathbf{c}_\text{i}$'s projection point on the line $\ell$. As  Fig.~\ref{fig:CD} shows, this averaging operation causes a small angle between $\mathbf{c}_\text{v}$ and $\mathbf{c}_\text{f-YIQ}$.
	
	The VIF averaging method in HSV color space averages $\mathbf{c}_\text{v}$ with $\mathbf{c}_\text{i}$ only in the $V$  channel (the length of the color vector)                                                                                                                                                                                                                                                                                                                                                                                                                               and keeps $H$ (hue) channel and $S$ (saturation) channel unchanged, so $\mathbf{c}_\text{v}$ and $\mathbf{c}_\text{f-HSV}$ are collinear. As  Fig.~\ref{fig:CD} shows, the angle between $\mathbf{c}_\text{v}$ and $\mathbf{c}_\text{f-HSV}$ is close to 0.
	
	Our method scales $\mathbf{c}_\text{v}$ by the scaling factor $k$, so $\mathbf{c}_\text{v}$ and the scaled vector $\mathbf{c}_\text{s}$ are collinear. The bounding operation limits the three channels of $\mathbf{c}_\text{s}$ to the range of [0,255], which may cause a tiny angle between $\mathbf{c}_\text{v}$ and $\mathbf{c}_\text{f-Ours}$, as shown in Fig.~\ref{fig:CD}.
	
	In general, given the same input images, the relationship of color deviation corresponding to these four methods is:
	\begin{equation}
		0\approx\text{CD}_\text{Ours}
		\approx\text{CD}_\text{HSV}
		<\text{CD}_\text{YIQ}
		<\text{CD}_\text{RGB}.
	\end{equation}
	Thus, our method (FCDFusion) and averaging in HSV color space better  preserve the hue and saturation of the visible image, a conclusion  confirmed by our experimental results in the next section.
	
	Unlike  averaging in HSV color space, our method does not need color space transformations and  embeds gamma correction to improve  contrast, so  is faster and can provide clearer objects in fused images.
	
	\section{Results}\label{sec:results}
	In this section, we compare our method to several state-of-the-art VIF methods. We first introduce the data set, evaluation metrics, and comparison methods used in the experiment, and then compare the fusion methods using metrics and visual effects.
	
	\subsection{Dataset}
	To ensure fairness and to reflect the processing of color information, we used visible and infrared image pairs from the visible and infrared image fusion benchmark (VIFB) \cite{VIFB} for real data experiments. VIFB is the only existing benchmark that provides image pairs with color visible images and provides unified procedures for state-of-art methods and evaluation metrics on the same computing platform. It provides 21 image pairs, a code library for 20 fusion algorithms, and 13 evaluation metrics. The average image size is $452\times 368$.

	VIFB not only provides visible and infrared image pairs of dim scenes but also provides some for backlight scenes, to ensure a comprehensive evaluation of fusion methods.
	
	\subsection{Metrics}
	We first use 13 existing VIF metrics provided by VIFB: CE \cite{CE}, EN \cite{EN}, MI \cite{MI}, AG \cite{AG}, EI \cite{EI}, SD \cite{SD}, SF \cite{SF}, $Q^\text{AB/F}$ \cite{QABF}, $Q_\text{CB}$ \cite{QCB}, $Q_\text{CV}$ \cite{QCV}, PSNR \cite{PSNR}, SSIM \cite{SSIM}, and RMSE \cite{PSNR}.
	As noted earlier, they mainly measure the contrast of the fused image itself and the similarity between the fused image and the two input images. 

    Since such existing metrics cannot fully assess the visual quality of the fused results, we invited 10 users to judge the relative quality of groups of fused images in VIFB (partly shown in Figs.~\ref{fig:shadow} and~\ref{fig:light}), focusing on object clarity and color authenticity. In each image group, the fused images obtained by 8 methods were randomly reordered and  scored from 1 to 8. Users can see the corresponding input images, but they do not know which method is used to obtain each output image.
    
	We also counted the number of FLOPs and parameters used by each method. The FLOPs  statistic measures the  number of equivalent floating-point multiplications and divisions required to fuse a pair of 452$\times$368 input images. The parameters statistic measures the number of parameters to be learn in neural networks or to be manually set in filters and other transformations.
 
	We also used the proposed color deviation (CD) metric to measure the color preservation abilities of the fusion methods.

	\subsection{Methods}
	We compared the proposed FCDFusion method to 3 simple averaging methods, the 2 highest-scoring methods from VIFB, and 2 state-of-the-art methods absent from VIFB.
	
	The 3 simple averaging methods using RGB, YIQ, and HSV color spaces are named RGB-AVG, YIQ-AVG, and HSV-AVG for short.
	
	The 2 highest-scoring methods from VIFB are MST-SR \cite{MST} and CNN \cite{CNN}. MST-SR is a hybrid method, which has 6 top-three metric values in VIFB, for CE, EN, MI, $Q^\text{AB/F}$, $Q_\text{CB}$ and $Q_\text{CV}$. CNN is a deep learning method based on convolutional neural networks, which has 5 top-three metric values in VIFB, for MI, $Q^\text{AB/F}$, $Q_\text{CB}$, $Q_\text{CV}$ and SD. These two methods were originally designed to fuse grayscale images. In VIFB \cite{VIFB}, they have been modified to fuse color images by fusing every channel of the RGB visible image with the corresponding infrared image.
	
	The 2 state-of-the-art methods absent from VIFB are PIAFusion \cite{PIAFusion} and SeAFusion \cite{SeAFusion}. They are based on deep learning and use YUV (or the equivalent YCrCb) \cite{YUV} color space.

	\begin{table*}[t!]
		\centering
		\caption{Average metric values for 8 methods on 21 image pairs provided by VIFB. The best three values for each metric are indicated in red, green, and blue, respectively. $\uparrow$  means that a larger value is better, while $\downarrow$ means that a smaller value is better.} 
		\begin{tabular}{l|ccc|cc|cc|c}
\hline
\textbf{Metric}          & \textbf{RGB-AVG}                & \textbf{YIQ-AVG}               & \textbf{HSV-AVG}                & \textbf{MST-SR}                & \textbf{CNN}                    & \textbf{PIAFusion}              & \textbf{SeAFusion}              & \textbf{FCDFusion (ours)}     \\ \hline
CE $\downarrow$          & 1.3335                          & 1.3844                         & 1.3639                          & {\color[HTML]{FF0000} 0.9572}  & {\color[HTML]{00B050} 1.0299}   & {\color[HTML]{0070C0} 1.2918}   & 1.5195                          & 1.5368                        \\
EN $\uparrow$            & 6.6814                          & 6.7032                         & 6.7311                          & {\color[HTML]{FF0000} 7.3391}  & {\color[HTML]{00B050} 7.3202}   & {\color[HTML]{0070C0} 6.9962}   & 6.9960                          & 6.8778                        \\
MI $\uparrow$            & 2.1976                          & 2.1444                         & 2.1233                          & {\color[HTML]{FF0000} 2.8090}  & {\color[HTML]{00B050} 2.6533}   & {\color[HTML]{0070C0} 2.4964}   & 2.1438                          & 2.2788                        \\
AG $\uparrow$            & 3.2779                          & 3.3192                         & 3.5214                          & {\color[HTML]{00B050} 5.8513}  & {\color[HTML]{0070C0} 5.8077}   & {\color[HTML]{FF0000} 5.8780}   & 5.6707                          & 3.9498                        \\
EI $\uparrow$            & 34.1354                         & 34.6171                        & 36.5583                         & {\color[HTML]{00B050} 60.7805} & {\color[HTML]{0070C0} 60.2406}  & {\color[HTML]{FF0000} 60.9310}  & 59.0180                         & 41.1680                       \\
SD $\uparrow$            & 34.1786                         & 34.8890                        & 34.9258                         & {\color[HTML]{00B050} 57.3134} & {\color[HTML]{FF0000} 60.0753}  & {\color[HTML]{0070C0} 52.3719}  & 50.0449                         & 41.5048                       \\
SF $\uparrow$            & 10.3152                         & 10.4263                        & 11.1832                         & {\color[HTML]{00B050} 18.8067} & {\color[HTML]{FF0000} 18.8130}  & {\color[HTML]{0070C0} 18.7830}  & 17.8480                         & 12.7951                       \\
$Q^\text{AB/F} \uparrow$ & 0.3995                          & 0.4037                         & 0.4177                          & {\color[HTML]{FF0000} 0.6611}  & {\color[HTML]{00B050} 0.6576}   & {\color[HTML]{0070C0} 0.6394}   & 0.5632                          & 0.4839                        \\
$Q_\text{CB} \uparrow$   & 0.4415                          & 0.4434                         & 0.4465                          & {\color[HTML]{FF0000} 0.6447}  & {\color[HTML]{00B050} 0.6215}   & 0.5400                          & 0.4627                          & {\color[HTML]{0070C0} 0.5428} \\
$Q_\text{CV} \downarrow$ & 747.9871                        & 749.6857                       & 767.9276                        & 522.6890                       & {\color[HTML]{0070C0} 512.5690} & {\color[HTML]{FF0000} 383.2480} & {\color[HTML]{00B050} 405.2917} & 774.8971                      \\ \hline
PSNR $\uparrow$          & {\color[HTML]{FF0000} 29.22745} & {\color[HTML]{00B050} 29.2022} & {\color[HTML]{0070C0} 29.18195} & 28.9754                        & 28.96605                        & 28.81665                        & 28.6775                         & 28.8974                       \\
SSIM $\uparrow$          & {\color[HTML]{FF0000} 0.7453}   & {\color[HTML]{00B050} 0.74205} & {\color[HTML]{0070C0} 0.7361}   & 0.69515                        & 0.69545                         & 0.6962                          & 0.69725                         & 0.71645                       \\
RMSE $\downarrow$        & {\color[HTML]{FF0000} 0.0516}   & {\color[HTML]{00B050} 0.05245} & {\color[HTML]{0070C0} 0.05295}  & 0.05825                        & 0.0589                          & 0.06195                         & 0.06625                         & 0.0586                        \\ \hline
User $\uparrow$          & 4.1833                          & {\color[HTML]{0070C0} 5.3167}  & {\color[HTML]{00B050} 5.6333}   & 3.2833                         & 3.3000                          & 3.8000                          & 3.2500                          & {\color[HTML]{FF0000} 7.2333} \\ \hline
Params $\downarrow$      & {\color[HTML]{FF0000} 0}        & {\color[HTML]{FF0000} 0}       & {\color[HTML]{FF0000} 0}        & {\color[HTML]{00B050} 84.40K}  & 435.20K                         & 1.18M                           & {\color[HTML]{0070C0} 166.66K}  & {\color[HTML]{FF0000} 0}      \\
FLOPs $\downarrow$       & {\color[HTML]{FF0000} 23.76K}   & 2.99M                          & {\color[HTML]{0070C0} 2.33M}    & 257.86M                        & 10.85G                          & 195.46G                         & 27.61G                          & {\color[HTML]{00B050} 1.16M}  \\ \hline
CD $\downarrow$          & 0.0656                          & {\color[HTML]{0070C0} 0.0411}  & {\color[HTML]{FF0000} 0.0111}   & 0.0711                         & 0.0669                          & 0.0366                          & 0.0436                          & {\color[HTML]{00B050} 0.0117} \\ \hline
\end{tabular}\label{tab:result}
	\end{table*}
	
	\subsection{Metric results}
	Results using contrast- and similarity-based metrics in Tab.~\ref{tab:result} show that our method provides higher contrast than the three simple averaging methods, and retains more feature details from the input images than MST-SR, CNN, PIAFusion, and SeAFusion. 
	
	MST-SR and CNN  score highly on contrast but destroy the color consistency with the original visible image, so have low scores for similarity. On the other hand, RGB-AVG  scores highly for similarity to the the two input images, by simply averaging them, but gets low scores for contrast because of the low saturation of the infrared image. Our method balances these two aspects, so it lies in the middle of the 8 methods when evaluated by the 13 existing metrics.

        Our user study shows that the images fused by our method have the best visual effects. Indeed, some complex methods based on multi-scale transformations (MST-SR) and deep learning (CNN, PIAFusion, and SeAFusion) are not even as effective as simple averaging methods (RGB-AVG, YIQ-AVG, and HSV-AVG) in this respect.
	
	The FLOPs metric shows the computational requirements of our method to be  much lower than for state-of-the-art methods such as MST-SR, CNN, PIAFusion, and SeAFusion. Among the fast methods, RGB-AVG is the fastest, but also the least effective when considering color and contrast. Our method provides higher contrast and better visual effects than YIQ-AVG and HSV-AVG, and has a much greater processing speed. At the same time, our method does not need memory to store parameters, so the requirements for computing devices are very low.
	
	The color deviation metric results show that our method and HSV-AVG  better retain the hue and saturation of the visible image. As noted in Sec.~\ref{sec:CDMotivation}, the final bounding operation (to a maximum of 255) in our method leads to bias; the bias angle of our method is slightly larger than that of HSV-AVG.
	
	\subsection{Visual effects}
\begin{figure*}[tp!]
		\centering
		\includegraphics[width=0.92\linewidth]{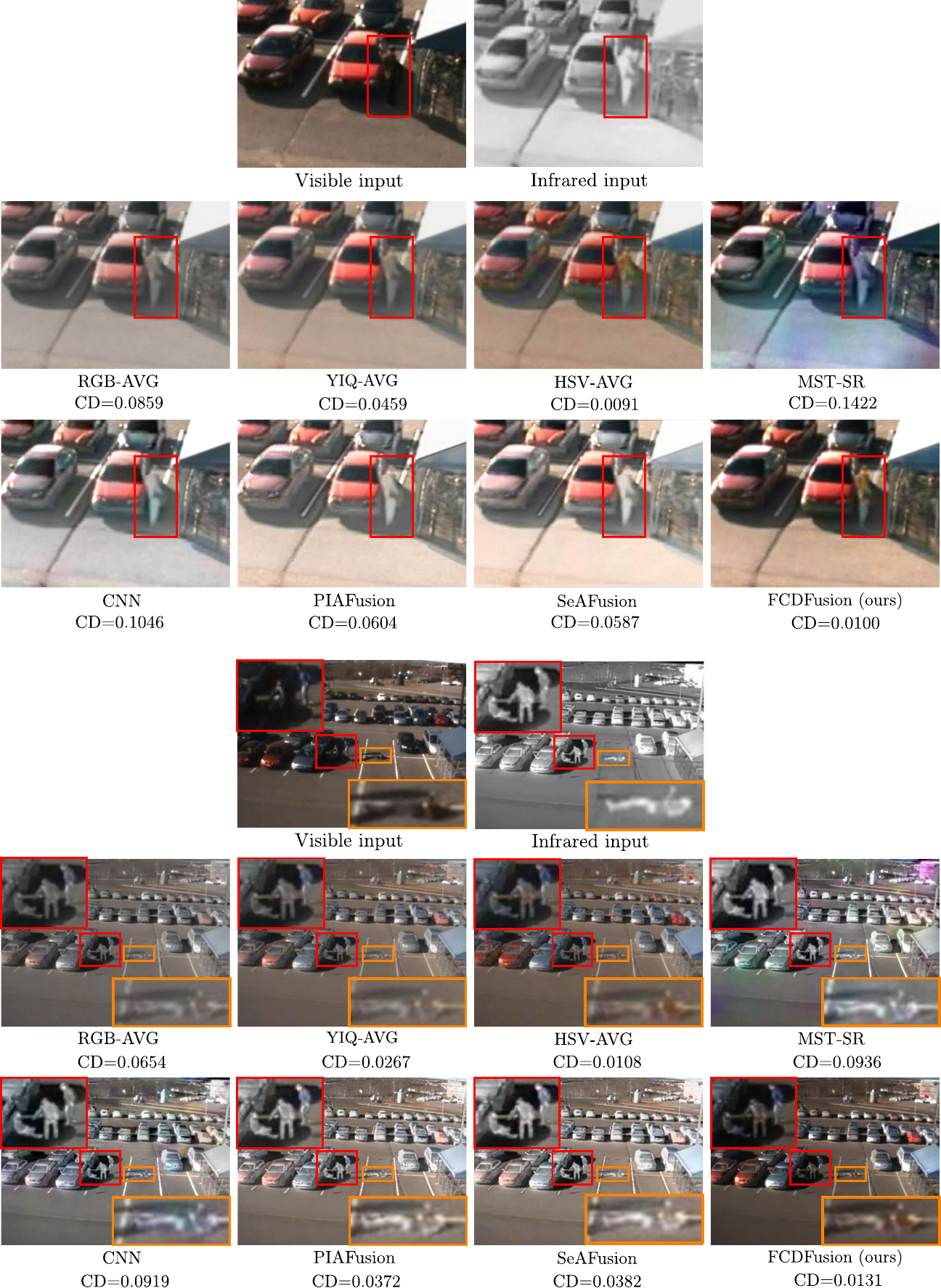}
		\caption{Fusion results showing objects in shadow. CD values give the corresponding  color deviation metric; lower is better.}
		\label{fig:shadow}
	\end{figure*}

\begin{figure*}[tp!]
		\centering
		\includegraphics[width=.92 \linewidth]{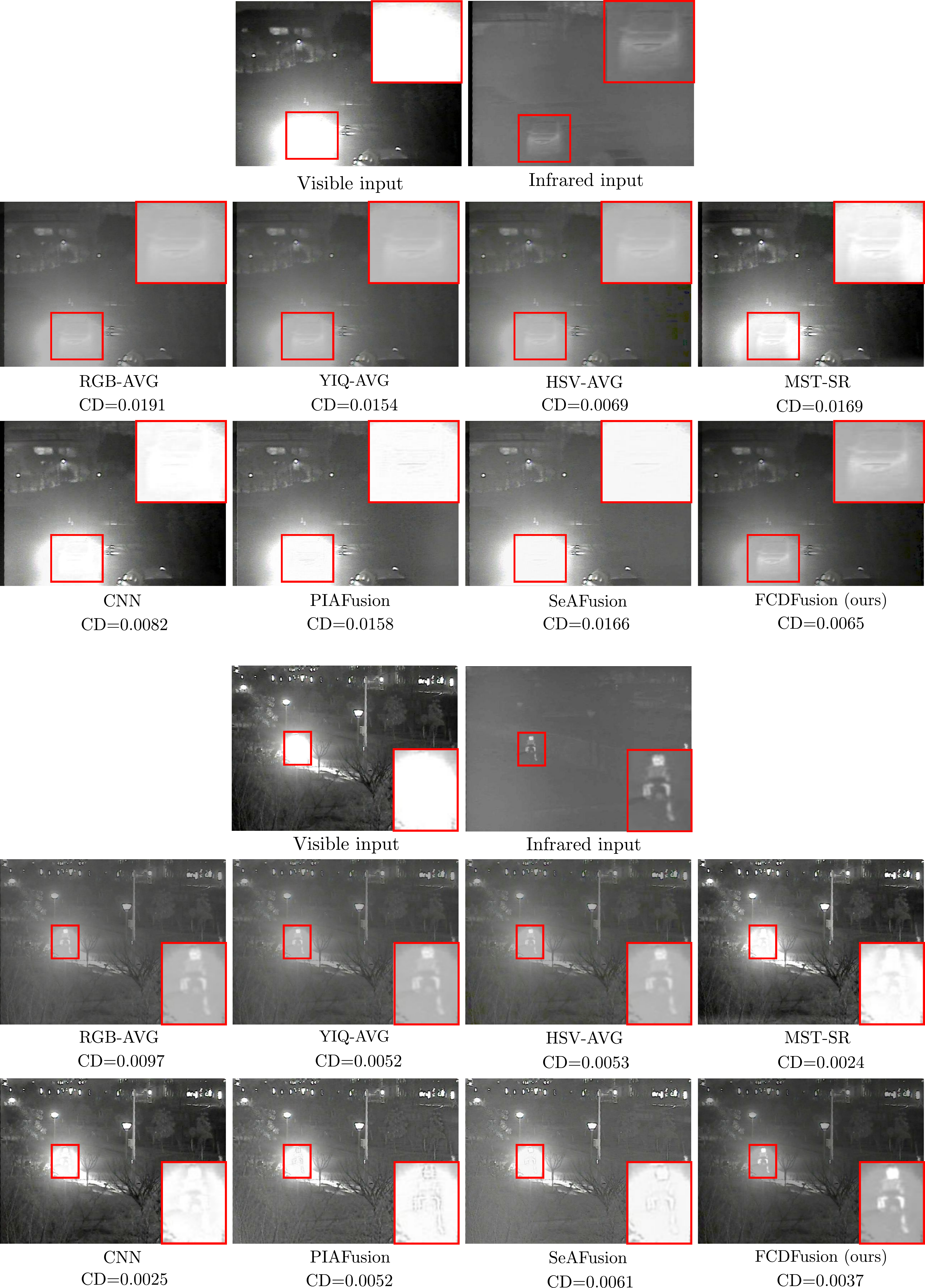}
		\caption{Fusion results showing objects in strong light. CD values give the corresponding  color deviation metric; lower is better.}
		\label{fig:light}
	\end{figure*}

	The groups of fused results in Figs.~\ref{fig:shadow},~\ref{fig:light} show two typical conditions found in VIF tasks, with objects in shadow and objects in strong light.
	
	Objects in shadow need enhancement by brightening. As shown in Fig.~\ref{fig:shadow},  images fused by our method have more contrast than that those from RGB-AVG, YIQ-AVG, and HSV-AVG,  making objects in shadows clearer. The reason is that our method uses gamma correction to improve contrast. Meanwhile, our method maintains the original color from the visible images, which avoids the loss of color information seen in the  images fused by RGB-AVG, YIQ-AVG, MST-SR, CNN, PIAFusion, and SeAFusion. Our method only changes the brightness of the color by vector scaling, and keeps its original hue and saturation.
	
	Objects in bright light need enhancement by dimming the light around them. As shown in Fig.~\ref{fig:light},  images fused by our method have more contrast than those from all other methods, making object in the bright light clearer, again due to the use of gamma correction. Deep learning-based methods (\emph{i.e.}, CNN, PIAFusion, and SeAFusion) do not perform under such conditions, partly because of the lack of training data for backlit scenes.
	
	Color deviation values indicated in Figs.~\ref{fig:shadow},~\ref{fig:light} are consistent with the color differences between the visible images and fused images as can be observed visually and intuitively. This confirms that the proposed color deviation metric is a practical way of measuring the color preservation abilities of VIF methods. The examples in Fig.~\ref{fig:light} have much smaller color deviation values than those in Fig.~\ref{fig:shadow}, as the visible images in Fig.~\ref{fig:light} are less colorful.
	
	\begin{figure*}[tp!]
		\centering
		\includegraphics[width=.72 \linewidth]{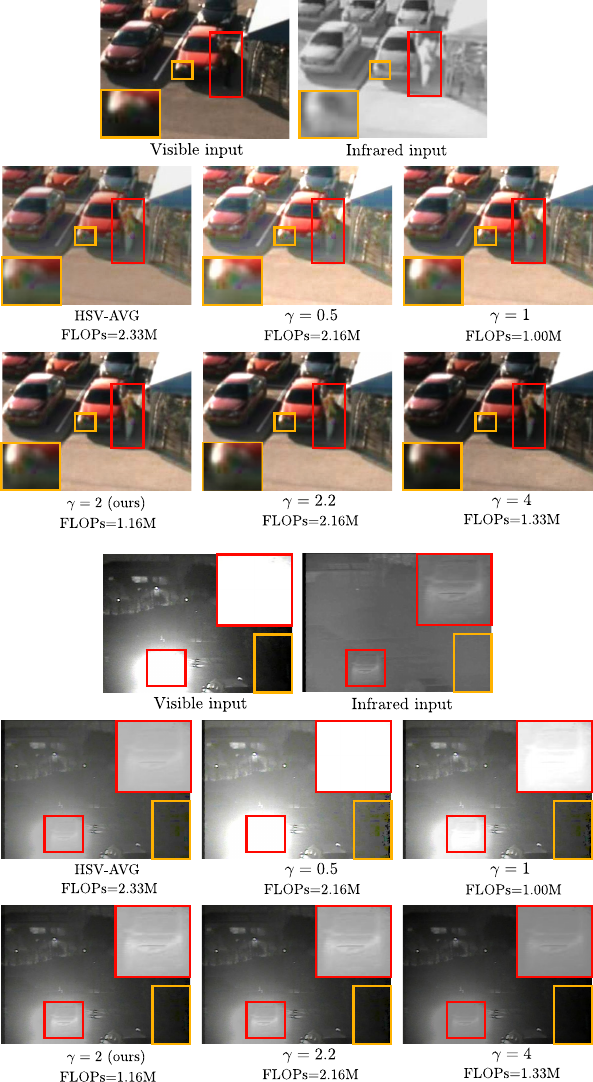}
		\caption{Fusion results from HSV-AVG, and our method using different gamma correction. Orange boxes highlight areas with color noise.}
		\label{fig:ablationGamma}
	\end{figure*}
	
	\begin{figure*}[t!]
		\centering
		\includegraphics[width=.95 \linewidth]{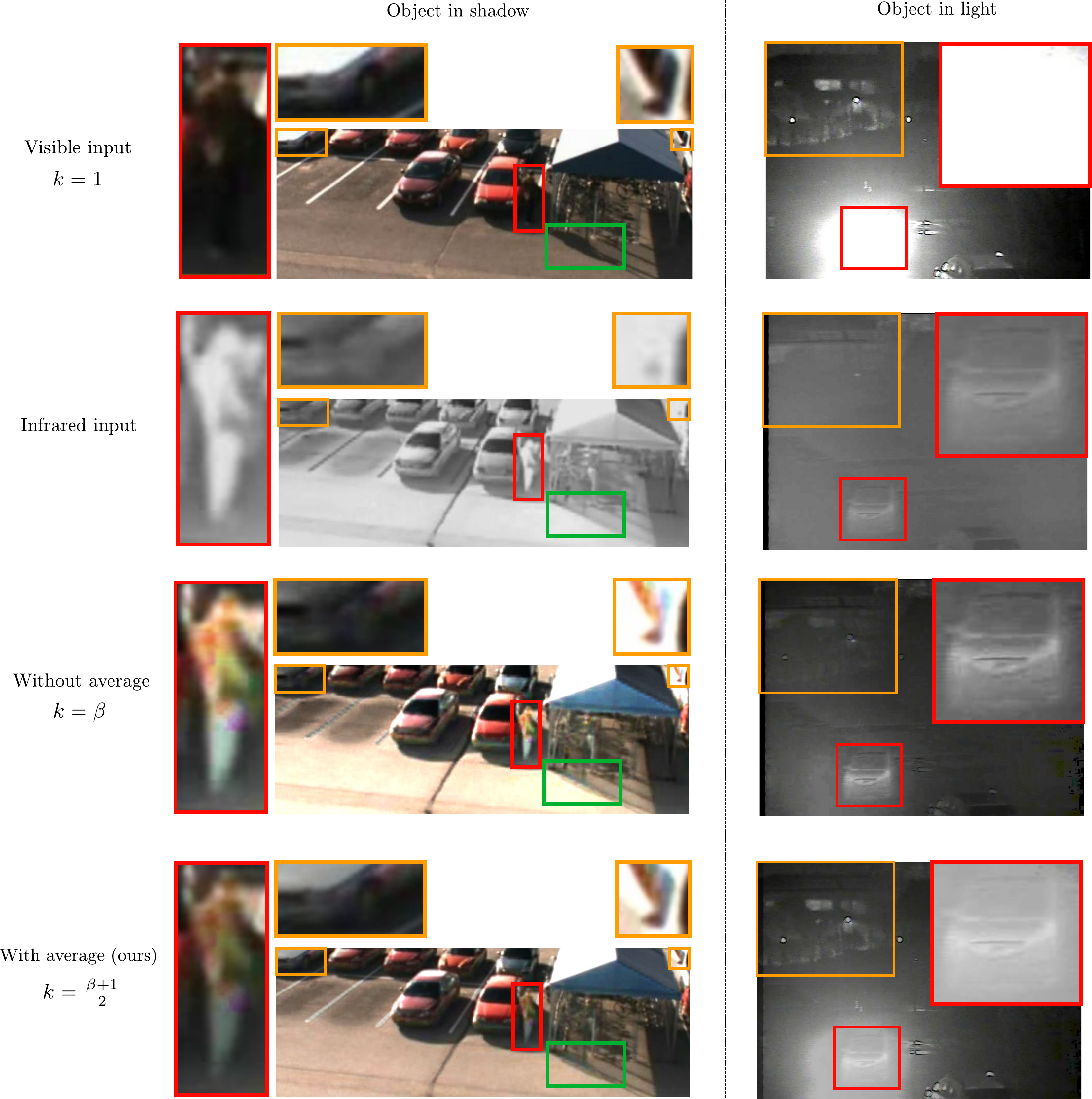}
		\caption{Fusion results with and without averaging when computing scaling factor $k$.}
		\label{fig:ablationAverage}
	\end{figure*}
	
	\section{Discussion}
	We now discuss the important roles of gamma correction and averaging in our method, which properly adjust the contrast of the fused image.
	
	\subsection{Gamma correction}
	Using gamma correction in our method enhances contrast and reduces color noise, as demonstrated  in Fig.~\ref{fig:ablationGamma}.
	
	Setting $\gamma=2$ makes the objects clearest in both shadow and light, while keeping the background  little changed from the visible image. Making $\gamma$  too small or too great will lighten or darken the background respectively, resulting in a decrease in background contrast. Setting $\gamma=1$ gives results like those of HSV-SVG.
	
	Furthermore, setting $\gamma=2$ suppresses color noise in  dark areas of the visible images. Using HSV-AVG, $\gamma=0.5$ or $\gamma=1$ lightens the dark areas in the visible images, such as the car's shadow in the first scene and the right area of the second scene. Thus color noise is magnified at the same time.
	
	By setting $\gamma=2$, the exponential operation in Eq.~(\ref{alpha}) is reduced to squaring, which can be computed much faster than when using a general value for gamma. The FLOPs values given in Fig.~\ref{fig:ablationGamma} show that our method is faster than HSV-AVG or when using other $\gamma$ settings (apart from $\gamma=1$). Gamma correction using $\gamma=2.2$ is widely used in display enhancement \cite{gamma}.  Fig.~\ref{fig:ablationGamma} shows that using $\gamma=2$ has very similar visual effects to using $\gamma=2.2$, so we set $\gamma=2$ for  speed.
	
	\subsection{Averaging}
	Average  $\beta$ and 1 when computing the scaling factor $k$ in our method helps to avoid excessive changes in brightness, as demonstrated in Fig.~\ref{fig:ablationAverage}. Without this averaging operation, objects are clear but some features in the background may become blurred, as the brightness information comes is mainly from the infrared image. The averaging operation provides brightness information from the visible image in the fused image. In applications that only focus on objects, the version without averaging may be better.
	
	\section{Conclusions}
	In this paper, we have proposed a simple and effective VIF method called FCDFusion, which preserves color information without color space transformations. It directly operates in RGB color space and  embeds gamma correction using little extra computation, so color and contrast are quickly improved. A theoretical analysis and experimental results show that our method can achieve satisfactory results in only 7 FLOPs per pixel. Compared to state-of-the-art fast and color-preserving methods using HSV color space, our method provides higher contrast and the computational cost is only half.
	
	In addition, we have proposed a new metric, color deviation, to measure the color preservation ability of a VIF method; it is specifically designed for VIF tasks using color visible images and overcomes the deficiencies of existing VIF metrics for color information evaluation.
	
	\appendix
	
	\subsection*{Acknowledgements}
    This work was supported by the National Natural Science Foundation of China under Grants Nos. 62171038, 61827901, and 62088101. We would like to thank all the authors who provided code to the visible and infrared image fusion benchmark.
	
	\subsection*{Declaration of competing interest}
	The authors have no competing interests to declare that are relevant to the content of this article.

	\bibliographystyle{CVMbib}
	\bibliography{main}

\begin{thebibliography}{10}
\expandafter\ifx\csname urlstyle\endcsname\relax
  \providecommand{\doi}[1]{doi:\discretionary{}{}{}#1}\else
  \providecommand{\doi}{doi:\discretionary{}{}{}\begingroup \urlstyle{rm}\Url}\fi

\bibitem{survey}
Ma J, Ma Y, Li C. Infrared and visible image fusion methods and applications: a survey. \emph{Information Fusion}, 2019, 45: 153--178.

\bibitem{detection}
Cao Y, Guan D, Huang W, Yang J, Cao Y, Qiao Y. Pedestrian detection with unsupervised multispectral feature learning using deep neural networks. \emph{Information Fusion}, 2019, 46: 206--217.

\bibitem{detect3}
Gao S, Cheng Y, Zhao Y. Method of visual and infrared fusion for moving object detection. \emph{Optics Letters}, 2013, 38(11): 1981--1983.

\bibitem{detect1}
Han J, Bhanu B. Fusion of color and infrared video for moving human detection. \emph{Pattern Recognition}, 2007, 40(6): 1771--1784.

\bibitem{detect2}
Ulusoy I, Yuruk H. New method for the fusion of complementary information from infrared and visual images for object detection. \emph{IET Image Processing}, 2011, 5(1): 36--48.

\bibitem{tracking}
Li C, Zhu C, Huang Y, Tang J, Wang L. Cross-modal ranking with soft consistency and noisy labels for robust RGB-T tracking. In \emph{Proceedings of the European Conference on Computer Vision}, 2018, 808--823.

\bibitem{track1}
Liu H, Sun F. Fusion tracking in color and infrared images using joint sparse representation. \emph{Science China Information Sciences}, 2012, 55(003): 590--599.

\bibitem{track2}
Smith D, Singh S. Approaches to multisensor data fusion in target tracking: a survey. \emph{IEEE Transactions on Knowledge and Data Engineering}, 2006, 18(12): 1696--1710.

\bibitem{recog1}
Hariharan H, Koschan A, Abidi B, Gribok A, Abidi M. Fusion of visible and infrared images using empirical mode decomposition to improve face recognition. In \emph{IEEE International Conference on Image Processing}, 2007, 2049--2052.

\bibitem{recog3}
Heo J, Kong SG, Abidi BR, Abidi MA. Fusion of visual and thermal signatures with eyeglass removal for robust face recognition. In \emph{IEEE/CVF Conference on Computer Vision and Pattern Recognition Workshops}, 2004, 122--122.

\bibitem{recog2}
Kong SG, Heo J, Abidi BR, Paik J, Abidi MA. Recent advances in visual and infrared face recognition - a review. \emph{Computer Vision and Image Understanding}, 2005, 97(1): 103--135.

\bibitem{surveil3}
Conaire C{\'O}, O'Connor NE, Cooke E, Smeaton AF. Comparison of fusion methods for thermo-visual surveillance tracking. In \emph{International Conference on Information Fusion}, 2006, 1--7.

\bibitem{surveil1}
Kumar P, Mittal A, Kumar P. Fusion of thermal infrared and visible spectrum video for robust surveillance. In \emph{Indian Conference on Computer Vision, Graphics \& Image Processing}, 2006, 528--539.

\bibitem{surveil2}
Simone G, Farina A, Morabito FC, Serpico SB, Bruzzone L. Image fusion techniques for remote sensing applications. \emph{Information Fusion}, 2002, 3(1): 3--15.

\bibitem{color-vision1}
Yin S, Cao L, Ling Y, Jin G. One color contrast enhanced infrared and visible image fusion method. \emph{Infrared Physics \& Technology}, 2010, 53(2): 146--150.

\bibitem{remote1}
Fu MY, Zhao C. Fusion of infrared and visible images based on the second generation curvelet transform. \emph{Journal of Infrared and Millimeter Waves}, 2009, 28(4): 254--258.

\bibitem{remote2}
Li H, Ding W, Cao X, Liu C. Image Registration and Fusion of Visible and Infrared Integrated Camera for Medium-Altitude Unmanned Aerial Vehicle Remote Sensing. \emph{Remote Sensing}, 2017, 9(5).

\bibitem{IHS}
Al-Wassai FA, Kalyankar N, Al-Zuky AA. The IHS transformations based image fusion. \emph{arXiv:1107.4396}, 2011.

\bibitem{PCA}
Patil U, Mudengudi U. Image fusion using hierarchical PCA. In \emph{International Conference on Image Information Processing}, 2011, 1--6.

\bibitem{LP}
Bao-Shu W. Multi-sensor image fusion based on improved laplacian pyramid transform. \emph{Acta Optica Sinica}, 2007, 27(9): 1605--1610.

\bibitem{wavelet}
Pajares G, {Manuel de la Cruz} J. A wavelet-based image fusion tutorial. \emph{Pattern Recognition}, 2004, 37(9): 1855--1872.

\bibitem{NC}
Xiao Y, Wang K. Image fusion algorithm using nonsubsampled contourlet transform. In \emph{MIPPR 2007: Multispectral Image Processing}, volume 6787, 2007, 435--443.

\bibitem{MSVD}
Naidu V. Image fusion technique using multi-resolution singular value decomposition. \emph{Defence Science Journal}, 2011, 61(5): 479--484.

\bibitem{LatLRR}
Li H, Wu XJ. Infrared and visible image fusion using latent low-rank representation. \emph{arXiv:1804.08992}, 2018.

\bibitem{DL2019}
Li H, Wu Xj, Durrani TS. Infrared and visible image fusion with ResNet and zero-phase component analysis. \emph{Infrared Physics \& Technology}, 2019, 102: 103039.

\bibitem{CNN}
Liu Y, Chen X, Cheng J, Peng H, Wang Z. Infrared and visible image fusion with convolutional neural networks. \emph{International Journal of Wavelets, Multiresolution and Information Processing}, 2018, 16(03): 1850018.

\bibitem{CUFD}
Xu H, Gong M, Tian X, Huang J, Ma J. CUFD: An encoder–decoder network for visible and infrared image fusion based on common and unique feature decomposition. \emph{Computer Vision and Image Understanding}, 2022, 218: 103407.

\bibitem{PIAFusion}
Tang L, Yuan J, Zhang H, Jiang X, Ma J. PIAFusion: A progressive infrared and visible image fusion network based on illumination aware. \emph{Information Fusion}, 2022, 83-84: 79--92.

\bibitem{SeAFusion}
Tang L, Yuan J, Ma J. Image fusion in the loop of high-level vision tasks: A semantic-aware real-time infrared and visible image fusion network. \emph{Information Fusion}, 2022, 82: 28--42.

\bibitem{hybrid2}
Wang Z, Gong C. A multi-faceted adaptive image fusion algorithm using a multi-wavelet-based matching measure in the PCNN domain. \emph{Applied Soft Computing}, 2017, 61: 1113--1124.

\bibitem{hybrid3}
Yin M, Duan P, Liu W, Liang X. A novel infrared and visible image fusion algorithm based on shift-invariant dual-tree complex shearlet transform and sparse representation. \emph{Neurocomputing}, 2017, 226: 182--191.

\bibitem{MST}
Liu Y, Liu S, Wang Z. A general framework for image fusion based on multi-scale transform and sparse representation. \emph{Information fusion}, 2015, 24: 147--164.

\bibitem{hybrid1}
Zhang X, Ma Y, Fan F, Zhang Y, Huang J. Infrared and visible image fusion via saliency analysis and local edge-preserving multi-scale decomposition. \emph{Journal of the Optical Society of America A}, 2017, 34(8): 1400--1410.

\bibitem{luo2022}
Luo Y, He K, Xu D, Yin W, Liu W. Infrared and visible image fusion based on visibility enhancement and hybrid multiscale decomposition. \emph{Optik}, 2022, 258: 168914.

\bibitem{yin2022}
Yin W, He K, Xu D, Luo Y, Gong J. Significant target analysis and detail preserving based infrared and visible image fusion. \emph{Infrared Physics \& Technology}, 2022, 121: 104041.

\bibitem{EN}
Roberts JW, Van~Aardt JA, Ahmed FB. Assessment of image fusion procedures using entropy, image quality, and multispectral classification. \emph{Journal of Applied Remote Sensing}, 2008, 2(1): 1--28.

\bibitem{AG}
Cui G, Feng H, Xu Z, Li Q, Chen Y. Detail preserved fusion of visible and infrared images using regional saliency extraction and multi-scale image decomposition. \emph{Optics Communications}, 2015, 341: 199--209.

\bibitem{SSIM}
Wang Z, Bovik AC, Sheikh HR, Simoncelli EP. Image quality assessment: from error visibility to structural similarity. \emph{IEEE transactions on image processing}, 2004, 13(4): 600--612.

\bibitem{PSNR}
Jagalingam P, Hegde AV. A review of quality metrics for fused image. In \emph{Aquatic Procedia}, 2015, 133--142.

\bibitem{enhance}
Mu Q, Wang X, Wei Y, Li Z. Low and non-uniform illumination color image enhancement using weighted guided image filtering. \emph{Computational Visual Media}, 2021, 7: 529--546.

\bibitem{enhance2}
Li P, Huang Y, Yao K. Multi-algorithm fusion of RGB and HSV color spaces for image enhancement. In \emph{Chinese Control Conference}, 2018, 9584--9589.

\bibitem{YIQ}
Schwarz MW, Cowan WB, Beatty JC. An experimental comparison of RGB, YIQ, LAB, HSV, and opponent colour models. \emph{ACM Transactions on Graphics}, 1987, 6(2): 123--158.

\bibitem{HSV}
Smith AR. Color gamut transform pairs. \emph{ACM Siggraph Computer Graphics}, 1978, 12(3): 12--19.

\bibitem{YUV}
Maller J. RGB and YUV Color. \emph{FXScript Reference}, 2003.

\bibitem{metrics}
Liu Z, Blasch E, Xue Z, Zhao J, Laganiere R, Wu W. Objective assessment of multiresolution image Fusion algorithms for context enhancement in night vision: a comparative study. \emph{IEEE Transactions on Pattern Analysis and Machine Intelligence}, 2012, 34(1): 94--109.

\bibitem{CE}
Bulanon D, Burks T, Alchanatis V. Image fusion of visible and thermal images for fruit detection. \emph{Biosystems engineering}, 2009, 103(1): 12--22.

\bibitem{MI}
Qu G, Zhang D, Yan P. Information measure for performance of image fusion. \emph{Electronics Letters}, 2002, 38(7): 313--315.

\bibitem{EI}
Balakrishnan R, Priya R. Hybrid multimodality medical image fusion technique for feature enhancement in medical diagnosis. \emph{International Journal of Engineering Science Invention}, 2018, 2(Special issue): 52--60.

\bibitem{SD}
Rao YJ. Review article: in-fibre bragg grating sensors. \emph{Measurement Science and Technology}, 1997, 8(4): 355--375.

\bibitem{SF}
Eskicioglu AM, Fisher PS. Image quality measures and their performance. \emph{IEEE Transactions on Communications}, 1995, 43(12): 2959--2965.

\bibitem{QABF}
Xydeas C, , Petrovic V. Objective image fusion performance measure. \emph{Military Technical Courier}, 2000, 56(4): 181--193.

\bibitem{QCB}
Chen Y, Blum RS. A new automated quality assessment algorithm for image fusion. \emph{Image and Vision Computing}, 2009, 27(10): 1421--1432.

\bibitem{QCV}
Chen H, Varshney PK. A human perception inspired quality metric for image fusion based on regional information. \emph{Information Fusion}, 2007, 8(2): 193--207.

\bibitem{gamma}
Poynton C. Digital video and HD: algorithms and interfaces, 2012: 315--354.

\bibitem{VIFB}
Zhang X, Ye P, Xiao G. VIFB: a visible and infrared image fusion benchmark. In \emph{IEEE/CVF Conference on Computer Vision and Pattern Recognition Workshops}, 2020, 468--478.

\end{thebibliography}
	
	\subsection*{Author biography}

    \begin{biography}[HesongLi]{Hesong Li} is a master's student at the School of Computer Science and Technology, Beijing Institute of Technology, , China. He received his bachelor's degree from the School of Science, Dalian Maritime University in 2021. His research interests include artificial intelligence and image processing. 
    	\end{biography}

	\begin{biography}[YingFu]{Ying Fu} received a B.S. degree in electronic engineering from Xidian University, Xi'an, China, in 2009, an M.S. degree in automation from Tsinghua University, Beijing, China, in 2012, and a Ph.D. degree in information science and technology from the University of Tokyo, Japan, in 2015. She is currently a Professor in the School of Computer Science and Technology, Beijing Institute of Technology. Her research interests include computer vision, image and video processing, and computational photography. 
		\end{biography}
	
\end{document}